# Low Rank Variation Dictionary and Inverse Projection Group Sparse Representation Model for Breast Tumor Classification

Xiaohui Yang[1], Xiaoying Jiang[1], Wenming Wu[1], Juan Zhang[2], Dan Long[2],
Funa Zhou[3], Yiming Xu[3]

1. Data Analysis Technology Lab, Institute of Applied Mathematics, Henan University, Kaifeng 475004, China
2. Zhejiang Cancer Hospital, Hangzhou 310022, China
3. School of Computer and Information Engineering, Henan University, Kaifeng 475004, China

**Abstract**: Sparse representation classification achieves good results by addressing recognition problem with sufficient training samples per subject. However, SRC performs not very well for small sample data. In this paper, an inverse-projection group sparse representation model is presented for breast tumor classification, which is based on constructing low-rank variation dictionary. The proposed low-rank variation dictionary tackles tumor recognition problem from the viewpoint of detecting and using variations in gene expression profiles of normal and patients, rather than directly using these samples. The inverse projection group sparsity representation model is constructed based on taking full using of exist samples and group effect of microarray gene data. Extensive experiments on public breast tumor microarray gene expression datasets demonstrate the proposed technique is competitive with state-of-the-art methods. The results of Breast-1, Breast-2 and Breast-3 databases are 80.81%, 91.49% and 100% respectively, which are better than the latest literature.

## 1 Introduction

Breast cancer is the second leading cancer among the women worldwide. It is regarded as ahetero geneous group of diseases with complex and distinctive underlying molecular pathogenesis [1]. Many of the new genomic analysis tools offer great promise for classifications of tumors based on variations in gene expression.

These techniques have been used to deep delineate tumor groups or to identify candidate genes for tumor prognosis and therapeutic targeting. As such problems can be viewed as classification ones, various classification methods have been applied to analyze or interpret gene expression data resulting from DNA microarrays [2-3]. However, have the characteristics of small samples (patients), high dimensions (thousands of genes) and high redundancy [4], which impose a challenge to tumor classification.

Microarray gene expression data-based tumor classification mainly consists of clustering [5] and classification [6]，Furey et al. [7] used support vector machine (SVM). Shi et al. [8] proposed an improved diagonal discriminant analysis with sparse constraint for tumor classification. The most common classifier algorithm used to classify medical data is J48 decision tree for breast tumor. The serious drawbacks of the decision tree algorithm are over fitting, complexity, cost. There are various methods such as Boosting/Bagging to ensemble various classifiers and to provide the efficient classification. Sparse representation-based classification (SRC) was introduced by Wright et al. [9] for face recognition, in which the training images are used as the dictionary to code an input testing samples as a sparse linear combination of them via l1-norm minimization. Many scholars use SRC for gene expression profiling data. Zheng et al. [10] made use of singular value decomposition to learn a dictionary and then classified gene expression data of tumor subtypes based on SRC. Gan et al. [12] improved and generalized [11] by adding a weighted matrix. However, The sparse representation based classification (SRC) performs not very well for small sample data. Yang et al. proposed an inverse projection based pseudo-full-space representation classification (PFSRC) for face recognition [13] by utilizing existing available samples rather constructing auxiliary training samples. The PFSRC focused on exploiting complementary information between training samples and test samples, for example, when a portion of an image is occluded, non-occluded region of the same image still contains useful information for identification. However, gene expression data do not have this complementarity. In the SRC and PFSRC based methods, the dictionary is constructed by all training samples. The existing methods are all based on the information gene to do the follow-up work directly.

Recently, a new kind of matrix recovery, namely robust principal component analysis (RPCA), RPCA is called Low rank, the RPCA proposed by Candes et al. can recover a low rank matrix from highly corrupted measurements [15]. RPCA is the most widely used method in the field of image processing [23], however, existing methods focus on the low-rank part, the application of RPCA is rare in tumor classification. Liu et al. propose a novel RPCA-based method for classifying tumor samples [14]. For a special biological process, the expression profiles of most of the genes are flat. All these genes are considered as non-differential expression. It is natural to treat these data of non-differentially expressed genes as approximately low rank. Only a small number of genes are relevant to a special biological process, so the data of these differentially expressed genes can be treated as sparse perturbation signals. RPCA is applied to extract a subset of genes associated with a special biological process, they still use the information gene to make direct follow-up recognition. For breast tumor, we should pay attention to the tumor lesion part and normal part.

Deep-learning based classification methods have been proved effective for recognition. It is noted that the success of deep-learning relies on big data, complex net structure and advanced hardware. In this paper, we do not consider the method of learning, but rather focus on an improvement from the mathematical model and dictionary. According to the characteristics and group sparsity, we propose a new inverse-projection group sparse representation based classification (IPGSR) and gives the model solution. We use new thinking, the proposed low-rank variation dictionary tackles tumor recognition problem from the viewpoint of detecting and using variations in gene expression profiles of normal and patients, rather than directly using these samples. We just focus on representation and classification, and adopt simple gene selection method.

The remainder of this paper is organized as follows. The remainder of this paper is organized as follows. The presented robust breast tumor classification based on low-rank variation dictionary and IPGSRC is stated in Section 2. Extensive experimental results on breast tumor gene expression datasets are shown in Section 3. Finally, section 4 concludes the paper.

**2 Methodology**

In this section, the methodology of the presented robust breast tumor classification is described. Firstly, we consider weighted group sparse representation of ideas, we put forward inverse projection group sparse representation model and give the model solution. Moreover, we give the constructed of the variation dictionary.

**2.1 Sparse representation based on training samples dictionary**

SRC assumes that each test sample can be linearly represented by sufficient training samples from the same category [9]. The projection way of SRC is that each test sample is projected into the corresponding training sample space.

Given sufficient training samples of the $i$-th object class, any test sample from the same class will approximately lie in the linear span of the training samples associated with object $i$. Denote the training samples of all $k$ classes as the matrix $X = [X_1, X_2, \cdots X_C] = [x_1^1, x_2^1, \cdots, x_n^C]$, where the sub-matrix $X_i = [x_1^i, x_2^i, \cdots, x_{n_i}^i] \in R^{d \times n_i}$ stacks the training samples of class $i$. Then, the linear representation of a testing sample $y$ can be rewritten in terms of all training samples as:

$$y = X\alpha_0 + z, \tag{1}$$

where $\alpha_0$ is a sparse vector whose entries are zeros except those associated with the $i$-th class, and $z \in R^d$ is a noise term with bounded energy $\|z\|_2 < \varepsilon$.

**2.2 Pseudo-full-space representation based on training samples dictionary**

SRC shows significant performance when there are enough available training samples per subject. However, face recognition often suffers from insufficient training samples. To tackle this problem, a novel classification technique is presented based on utilizing existing available samples rather than constructing auxiliary training samples. PFSR is stably and effectively exploit complementary information between samples.

PFSR aims to seek a representation space as large as possible. The training and test sample space are *X* and *Y*, respectively. If the space { *X* , *Y* } is called full space, which contains all training samples and test samples. And then the space $V_j$ is the largest representation space of a sample $x_j$,

$$V_j = \{X,Y\} - x_j.$$

It is quite natural that the $V_j$ is just the full space except the training sample $x_j$ itself and called the pseudo-full-space of $x_j$. $V_j$ Obviously, provides richer information than the training sample space because of the addition of the test samples.

PFSR means that a training sample $x_j$ from a category *i* is represented by its corresponding pseudo-full-space.

$$x_j = \cdots + \beta_{i,j-1}x_{j-1} + \beta_{i,j+1}x_{j+1} \cdots + \gamma_{j,1}y_1 + \cdots + \gamma_{j,k}y_k. \qquad (2)$$

where $\beta_{i,s} \in R$ and $\gamma_{j,l} \in R$ are the corresponding coefficients before training samples and test samples respectively, $i = 1, 2, \cdots, C$, $j = 1, 2, \cdots, s_C$, $l = 1, 2, \cdots, k$.

Let $A_j = [\cdots, \beta_{i,j-1}, \beta_{i,j+1}, \cdots, \gamma_{j,1}, \cdots, \gamma_{j,k}]^T$, $s_{x_j}^i$ can be rewritten as

$$x_j = A_j V_j. \qquad (3)$$

**2.3 IPGSRC based on low-rank variation dictionary**

**2.3.1 Low rank model for gene expression data**

For SR and PFSR method, they all use information genes as features for classification. We are more concerned with the lesion part of the gene, in this paper, we use the low-rank to construct a new dictionary. Furthermore, the new variation dictionary has much sparse than the training samples. The proposed method tackles tumor recognition problem from the viewpoint of detecting and using variations in gene expression profiles of normal and patients, rather than directly using these samples as conditional.

Considering the matrix *X* of gene expression data with size $m \times n$, each row of *X* represents the transcriptional responses of a gene in all the *n* samples, and each column of *X* represents the expression levels of all the *m* genes in one sample.

Supposing that $X$ is given by $X = L_X + S_X$, $S_X = [S_{X_1}, S_{X_2}, \cdots, S_{X_C}] = [s_{x_1}^1, s_{x_2}^1, \cdots, s_{x_n}^C]$.
Low rank solves the following optimization problem:

$$\min_{L_X, S_X} \| L_X \|_* + \lambda \| S_X \|_1,$$
$$s.t. \ X = L_X + S_X. \tag{4}$$

where $\lambda$ is a positive regulation parameter, $\| L_X \|_* := \sum_i \sigma_i(L_X)$ denotes the nuclear norm of the matrix $X$, that is, the sum of its singular values, and $\| S_X \|_1 := \sum_{ij} | S_{X_{ij}} |$ denotes the $l_1$-norm of $S_X$.

A Lagrange multiplier $\Phi$ is introduced to remove the equality constraint of the Low rank problem in Eq.(4). According to [15], the augmented Lagrange multiplier method can be applied on the Lagrangian function:

$$L(L_X, S_X, \Phi, \mu) = \| L_X \|_* + \lambda \| S_X \|_1 + \langle Y, X - L_X - S_X \rangle + \frac{\mu}{2} \| X - L_X - S_X \|_F^2,$$

where $\rho$ is a positive scalar and $\| . \|_F^2$ denotes the Frobenius norm.

**2.3.2 Variation dictionary by low rank (Fixed dictionary)**

Low rank was originally proposed by Candes et al.[15], they goal of using low rank to model gene expression data is to classify tumor samples based on the characteristic genes that are identified by our method. However, they direct use of information genes for subsequent identification, we are concerned about the changes in the breast tumor, RPCA can decompose the observation matrix $L_X$ and give the sparse perturbation matrix $S_X = [S_{X_1}, S_{X_2}, \cdots, S_{X_C}] = [s_{x_1}^1, s_{x_2}^1, \cdots, s_{x_n}^C]$. The genes corresponding to non-zero entries in can be considered as ones of differential expression. Sparse perturbation matrix $S_X$ is breast tumor variations in gene expression profiles of normal and patients. So we construction of low-rank variation dictionary by sparse perturbation signals.

**2.3.3 Low-rank variation dictionary (Changing dictionary)**

For other problems, such as: to determine whether there is a breast cancer and the type of tumor, etc., in view of these problems, we construct a variable variation dictionary, according to the training sample changes the variation dictionary. For the training samples, because of the known class labels, low-rank decomposition can be performed with different types of samples (different classes can reflect the difference of the samples to the sparse parts of the low-rank decomposition). For the test sample, the test sample is low rank decomposed with the help of the existing training sample (the training sample contains both the same type and the different type of the test sample), because it does not know the class. The algorithm for constructing a variation dictionary is similar to a fixed variation dictionary, we give the algorithm flow of the variation dictionaries.

**Algorithm 1:** Variation dictionaries are constructed using the sparse parts of the low-rank decomposition

**Input**: Training sample set $X = [X_1, X_2, \cdots X_C] = [x_1^1, x_2^1, \cdots, x_n^C]$, and test sample set $Y = [y_1, y_2, \cdots, y_k]$. $k$ expresses the number of test samples.

Step 1: Given an appropriate parameter $\lambda = 1/\sqrt{\max(m,n)}$, where $m$ the transcriptional responses of a gene samples, $n$ represents the expression levels of all the $m$ genes in one sample.

Step 2: $X = L_X + S_X$, $Y = L_Y + S_Y$, where $L_X, L_Y$ denote low-rank matrix, $S_X = [S_{X_1}, S_{X_2}, \cdots, S_{X_C}] = [s_{x_1}^1, s_{x_2}^1, \cdots, s_{x_n}^C]$, $S_Y = [s_{y_1}, s_{y_2}, \cdots, s_{y_k}]$ denote sparse perturbation signals.

**Output**: variation dictionary $S_X = [S_{X_1}, S_{X_2}, \cdots, S_{X_C}] = [s_{x_1}^1, s_{x_2}^1, \cdots, s_{x_n}^C]$.

## 2.3.4 IPGSR based on low-rank variation dictionary

The PFSRC focused on exploiting complementary information between training samples and test samples, for example, when a portion of an image is occluded, non-occluded region of the same image still contains useful information for identification. However, gene expression profiling data do not have this complementarity with face, so we use the test sample space represent the training samples, The new presentation space is called inverse projection representation. It is known that encoding the group information in addition to sparsity will lead to better signal recovery/feature selection. The $l_{2,1}$-regularization promotes group sparsity. Combination of the group sparsity and the inverse projection representation, we proposed to inverse-projection group sparse representation (IPGSR).

IPGSR uses all the test samples to represent each training sample. The training and test sample space are $S_X$ and $S_Y$, the linear representation of a testing sample $s_{x_j}^i$ can be rewritten in terms of all variation dictionary as:

$$s_{x_j}^i = m_{i,1} s_{y_1} + \cdots + m_{i,1} s_{y_r} \cdots + m_{i,k} s_{y_k}. \qquad (5)$$

where $m_{i,r} \in R$. Let $m_i = [m_{i,1}, \ldots, m_{i,k}]^T$, $s_{x_j}^i$ can be rewritten as $s_{x_j}^i = S_Y m_i$. And then all training samples can be linearly represented as follows.

$$S_X = S_Y M, \qquad (6)$$

where $M = [M_{G_1}, \cdots, M_{G_c}] = [m_1, \cdots, m_n]^T \in R^{k \times n}$ is the projection coefficient matrix, $S_X = [S_{X_1}, S_{X_2}, \cdots, S_{X_C}]$ variation dictionary.

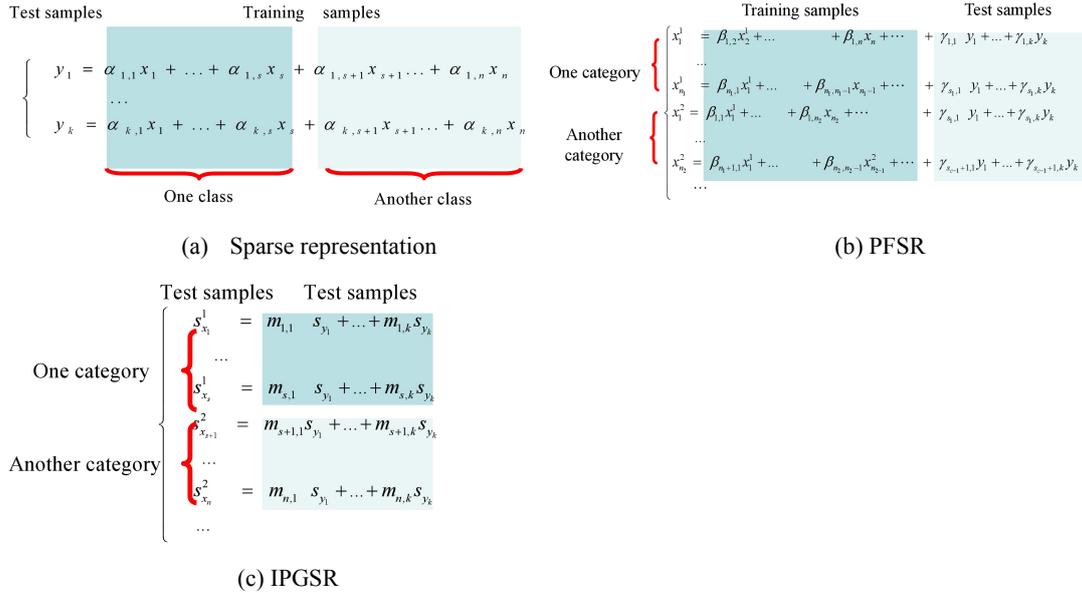

Fig.1. Comparison of representations. (a) SR based on training samples dictionary; (b)PFSR based on training samples dictionary; (c) IPGSR based on low-rank variation dictionary.

Comparing Eq. (2) and Eq. (5), IPGSR space has changed, comparing Eq. (1) and eq. (5), IPGSR is just the opposite of the SR representation space.

**2.3.5 Optimization of IPGSR model**

The group sparsity in SR is done by vector, we consider the group sparsity of the vector of variable selection. Therefore, the group sparse of the inverse projection representation model is to do the matrix.

The (weighted) $l_{w,2,1}$-regularization has been popularly used to encode the group sparsity [5], given by

$$\min_{M} \|M\|_{w,2,1} = \sum_{j=1}^{s} w_j \|M_{G_j}\|_F^2.$$

$$s.t. \quad S_Y M = S_X.$$

(7)

where $S_X \in R^{m \times n}$, $M = [M_{G_1}, \cdots, M_{G_c}] = [m_1, \cdots, m_n]^T \in R^{k \times n}$, $S_Y \in R^{m \times k}$,

$\|M\|_{21} = \sum_{i=1}^{n} \sqrt{\sum_{j=1}^{m} m_{ij}^2} = \sum_{i=1}^{n} \|m^i\|_2$. $m^i, m_j$ represent its $i$-th row and $j$-th column respectively, $\|m^i\|_2$ denotes the $l_2$-norm of $m^i$. $m_i, i=1,\cdots,n$ is coefficient vector. $G_p = p, p = 1,\cdots,c$, denotes the number of categories.

The optimization problem can be solved by the ADMM [17]. The problem (7) can be converted into an equivalent optimization problem as

$$\min_{M,Z} \|Z\|_{w,F,1} = \sum_{j=1}^{s} w_j \|Z_{G_j}\|_F.$$

$$s.t. \quad S_Y M = S_X, \quad Z = M.$$

(8)

Note that problem (8) has two blocks of variables ($M$ and $Z$) and its objective function is separable in the form of $f(M) + g(Z)$ since it only involves $Z$, thus ADDM is applicable. The augmented Lagrangian function of problem is defined by,

$$L_{\beta_1,\beta_2}(M,Z;\Lambda_1,\Lambda_2) = \|Z\|_{w,2,1} - \langle \Lambda_1, Z - M \rangle + \frac{\beta_1}{2}\|Z - M\|_F^2$$
$$- \langle \Lambda_2, S_Y M - S_X \rangle + \frac{\beta_2}{2}\|S_Y M - S_X\|_F^2.$$

where $\Lambda_1 \in R^{m \times n}, \Lambda_2 \in R^{m \times n}$, are multipliers and $\beta_1, \beta_2$ are penalty parameters.

The ADMM scheme takes the following form

$$\begin{cases} M \leftarrow \arg\min_{A} L_{\beta_1,\beta_2}(M,Z;\Lambda_1,\Lambda_2), \\ Z \leftarrow \arg\min_{Z} L_{\beta_1,\beta_2}(M,Z;\Lambda_1,\Lambda_2), \\ \Lambda_1 \leftarrow \Lambda_1 - \beta_1(Z - M), \\ \Lambda_1 \leftarrow \Lambda_2 - \beta_2(S_Y M - S_X). \end{cases}$$

Finally, we get the iteration formula

$$M \leftarrow (\beta_1 I + \beta_2 S_{Y^T} S_Y)^{-1}(-(\Lambda_1)^T + \beta_1 Z + (\Lambda_2)^T S_Y + \beta_2 S_{Y^T} S_X).$$

$$\min_{M,Z} \|Z\|_{w,F,1} - \Lambda_1^T(Z-M) + \frac{\beta_1}{2}\|Z-M\|_F^2 - \Lambda_2^T(YM-X) + \frac{\beta_2}{2}\|YM-X\|_F^2. \quad (9)$$

For $M$,

$$\Lambda_1^T I - \beta_1(Z-M) - \Lambda_2^T Y + \beta_2 Y^T(YM-X) = 0.$$

$$\Lambda_1^T I - \beta_1 Z + \beta_1 M - \Lambda_2^T Y + \beta_2 Y^T YM - \beta_2 Y^T X = 0.$$

$$(\beta_1 I + \beta_2 Y^T Y)M = -\Lambda_1^T + \beta_1 Z + \Lambda_2^T Y + \beta_2 Y^T X.$$

$$M \leftarrow (\beta_1 I + \beta_2 Y^T Y)^{-1}(-\Lambda_1^T + \beta_1 Z + \Lambda_2^T Y + \beta_2 Y^T X). \quad (10)$$

Minimizing (9) with respect to $Z$ gives the following $Z$ sub-problem:

Simple manipulation shows that (9) is equivalent to

$$\min_{Z} \sum_{i=1}^{s} [w_i \|Z_{G_i}\|_2 + \frac{\beta_1}{2}\|Z_{G_i} - M_{G_i} - \frac{1}{\beta_1}(\lambda_1)_{G_i}\|_2^2].$$

which has a closed form solution by the one-dimensional shrinkage (or soft thresholding) formula:

$$Z_{G_j} = \max\left\{\|\gamma_j\|_F - \frac{w_j}{\beta_1}, 0\right\} \frac{\gamma_j}{\|\gamma_j\|_F}. \quad (11)$$

where

$$\gamma_j = M_{G_j} + \frac{1}{\beta_1}(\Lambda_1)_{G_j}.$$

and the convention $0 \cdot \frac{0}{0} = 0$ is followed. We let the above group-wise shrinkage operation be denoted by $Z = shrink(M + \frac{1}{\beta_1}\Lambda_1, \frac{w}{\beta_1})$. for short.

Finally, the multipliers $\Lambda_1$ and $\Lambda_2$ are updated in the standard way

$$\Lambda_1 \leftarrow \Lambda_1 - \gamma_1 \beta_1(Z-M).$$
$$\Lambda_2 \leftarrow \Lambda_2 - \gamma_2 \beta_2(S_Y M - S_X).$$

where $\gamma_1, \gamma_2 > 0$ are step lengths.

In short, we have derived an ADDM iteration scheme for (8) as follows:

**Algorithm2**: ADMM for IPGSR
---
**Input:** Training sample set $S_X = [S_{X_1}, S_{X_2}, \cdots, S_{X_C}] = [s_{x_1}^1, s_{x_2}^1, \cdots, s_{x_n}^C]$, test sample set $S_Y = [s_{y_1}, s_{y_2}, \cdots, s_{y_k}]$.

**Initialize:** Initialize $\Lambda_1 \in R^{n \times k}, \Lambda_2 \in R^{m \times n}$, $M^0 = Z^0 = 0$, $\gamma_1, \gamma_2 > 0$, $\beta_1, \beta_2 > 0$.

Iterate until convergence:

1) $M \leftarrow (\beta_1 I + \beta_2 S_{Y^T} S_Y)^{-1}(-\Lambda_1^T + \beta_1 Z + \Lambda_2^T S_Y + \beta_2 S_{Y^T} S_X)$;

2) $Z_{G_j} \leftarrow \max\left\{\|\gamma_j\|_F - \frac{w_j}{\beta_1}, 0\right\} \frac{\gamma_j}{\|\gamma_j\|_F}$, $\gamma_j = M_{G_j} + \frac{1}{\beta_1}(\Lambda_1)_{G_j}$;

3) $\Lambda_1 \leftarrow \Lambda_1 - \gamma_1 \beta_1 (Z - M)$;

4) $\Lambda_2 \leftarrow \Lambda_2 - \gamma_2 \beta_2 (S_Y M - S_X)$.

End while
**Output:** An optimal solution $M, Z$.

---

### 2.3.6 Category contribution rate

From Figs.1, it can be also observed that the conventional classification criteria, reconstruction error, doesn't work for IPGSR. Since the representation dictionary is unlabeled test samples. Hence, category contribution rate (CCR) is constructed to match the proposed PFSRC [13] and complete classification. IPGSR using classification criteria of CCR.

For a test sample $s_{y_r}$, the contribution rate $C_{j,r}$ of $s_{y_r}$ for the $j$-th category can be calculated by Eq. (12)

$$C_{j,r} = \frac{1}{s_j} \sum \left( \frac{\delta_j(\{|m_{i,r}|\}_{i=1,\ldots,n})}{\sum_{i=1,\ldots,n} \{|m_{i,r}|\}_{i=1,\ldots,n}} \right). \tag{12}$$

where $j = 1, 2, \ldots, c, r = 1, 2, \ldots, k$, $s_j$ denotes the number of $j$-th category training samples.

The CCR matrix $[C_{j,r}], j = 1, 2, \ldots, c, r = 1, 2, \ldots, k$, for all test samples is got. Through the CCR, we can compare correlations between each test sample and every category. The larger the CCR is, the higher the correlation is. A test sample $s_{y_r}$ is classified into the category with the maximal contribution rate $\{C_{j,r}, j=1,2,\ldots,c\}$.

$$u_r = \arg\max_{j\in\{1,\cdots,c\}} (C_{j,r}). \tag{13}$$

By this means, categories of all test samples are obtained simultaneously and classification can be completed.

**2.3.7 Classification stability index**

For the representation-based classification methods, suppose $R_{best}^1$ and $R_{best}^2$ are the values of a classification criterion corresponding to the best category and the second best category. The CSI of a test samples is defined to measure the difference between $R_{best}^1$ and $R_{best}^2$. The CSI is normalized as $CSI \in [0,1]$ and is always defined as the ratio of the smaller one and the larger one.

$$CSI = \frac{R_{best}^1}{R_{best}^2}. \tag{14}$$

For SRC, the CSI is denoted as $CSI_{RE}$, where $R_{best}^1$ and $R_{best}^2$ are the minimal reconstruction error and the second minim alone. While for IPRC, the CSI is denoted as $CSI_{CCR}$, where $R_{best}^1$ and $R_{best}^2$ are the second maximal CCR and the maximal one.

By this way, a statistical measure, CSI, is defined qualify the classification stability of representation‐based methods. The smaller the index is, the better the stability is, the better the representation‐based method is. Detailed experiments will be shown in Subsection 3.3.4.

**2.4 Low-rank variation dictionary and IPGSRC model for breast tumor**

Combined the low-rank variation dictionary with IPGSRC, the basic idea of our robust breast tumor classification algorithm is as follows.

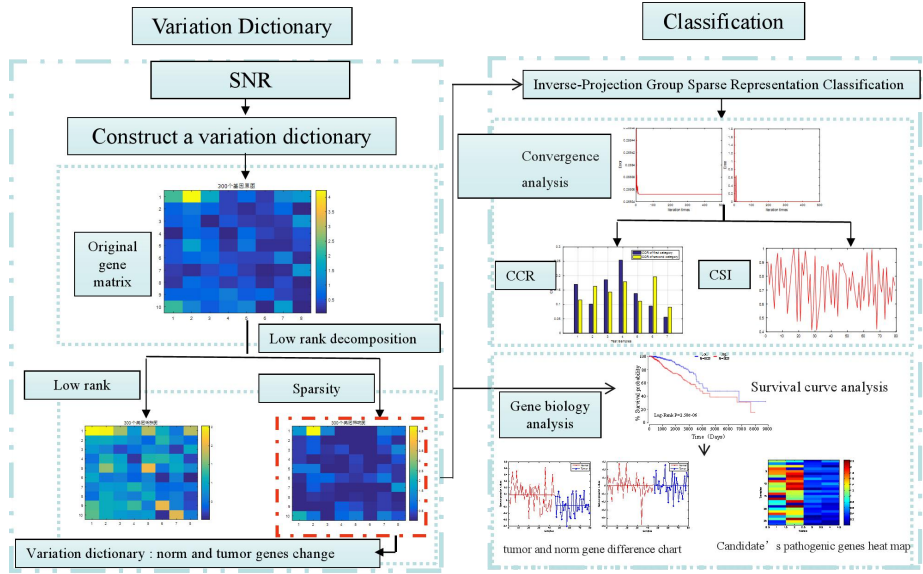

Fig. 2. Flow chart of the low rank variation dictionary and IPGSRC for robust breast tumor.

**Algorithm3**: Classification based on IPGSR.

Input: Low rank variation dictionary $S_X = [S_{X_1}, S_{X_2}, \cdots, S_{X_C}] = [s_{x_1}^1, s_{x_2}^1, \cdots, s_{x_n}^C]$ the test sample set $S_Y = [s_{y_1}, s_{y_2}, \cdots, s_{y_k}]$.

Step1. By Eq. (5) the inverse projection representation is realized.
Step2. By Eq. (10), the projection coefficient matrix is got.
Step3. By normalizing the CCR matrix, relevancies between each test sample and all categories are obtained.
Output: By Eq. (13), each test sample can be classified into the category with the maximal CCR.

## 3 Experiments and discussions

In this section, effectiveness of our methods is demonstrated by extensive experiments on public breast tumors gene expression datasets. The kinds of measures are used to measure the performance of these methods. Accuracy measures the classification performance by using the percentage of correctly classified samples. Sensitivity measures the non-missed diagnosis performance by using the rate of correctly classified positive samples. Specificity measures the non-misdiagnosis performance by using the rate of correctly classified negative samples. For any test, there is usually a trade-off between the sensitivity and specificity. This trade off can be represented graphically using a receiver operating characteristic curve (ROC), which is a graphical plot that illustrates the diagnostic ability of a binary classifier system as its discrimination threshold is varied[18]. AUC is just the area under the curve of ROC and is also suitable to binary classification problem. Error reduction rate (ERR)[19] intuitively characterizes the proportion of the errors reduced by switching a method to the other one. Without loss of generality, ten-fold cross-validation ten times is used to test the performance of the algorithms. The algorithm gives the convergence analysis and gene biological analysis.

All experiments are carried out using MATLAB R2016a on a 3.30GHz machine with 4.00GB RAM.

**3.1 Breast tumor datasets**

This breast tumor [20] [21]study was first reported in (Van't Veer et al,2002). This DNA microarray analysis on primary breast tumor of 25,000 gene expression Measurements of 117 young patients. We selected 79 primary breast tumor: 34 from patients who developed distant metastases within 5 years, 45 from patients who continued to be disease-free after a period of at least 5 years, Breast-2: all 79 patients were lymph node-negative, and 55 years old or younger. From each patient, tumor size under 5 cm.

Breast-2(97) [25]: For the samples of 117 databases Van't Veer et al[21], about 25,000 genes. Many articles select samples for further analysis and classification based on different indicators. Jiang et al. [25] Where 97 lymph node-negative breast tumor patients, 55 years old or younger, participated in this study. Among them, 46 developed distant metastases within 5 years and 51 remained metastases free for at least 5 years, which is reported as (Breast-2(97)).

Breast-1[24]: The purpose of the study is to classify female breast tumor patients according to relapse and no relapse clinical outcomes using gene expression data. Fan et al.[24] chose 77 samples on the basis of Van't Veer et al [21] .There were 44 developed distant metastases within 5 years and 33 remained metastases free for at least 5 years, which is reported as (Breast-1).

Breast-3: Molecular Signature of Pregnancy Associated Breast Cancer (PABC), Malignant epithelia and tumor-associated stroma of PABC and Non-PABC were isolated by laser capture microdissection and gene expression profiled. Additionally, normal breast epithelia and stroma adjacent to the two tumor types were profiled. Series_supplementary_fileftp://ftp.ncbi.nlm.nih.gov/geo/series/GSE31nnn/GSE31192

/suppl/GSE31192_RAW.tar.

For gene expression data, there are several simple ways to deal with missing values such as deletingan expression vector with missing values from further analysis, imputing missing values to zero, or imputing missing values of a certain gene (sample) to the sample(gene) average (Alizadeh et al., 2000). In this article, we used this method.

Table.1 Breast dataset

| Data sets | Class1 | Class2 | All | Genes |
| --- | --- | --- | --- | --- |
| Breast-1 | 33 | 44 | 78 | 4869 |
| Breast-2 | 45（norm） | 33(tumor) | 79 | 21220 |
| Breast-3 | 13（norm） | 20(tumor) | 33 | 54675 |

**3.2 Performance of variation dictionary**

**3.2.1 Sparsity of variation dictionary**

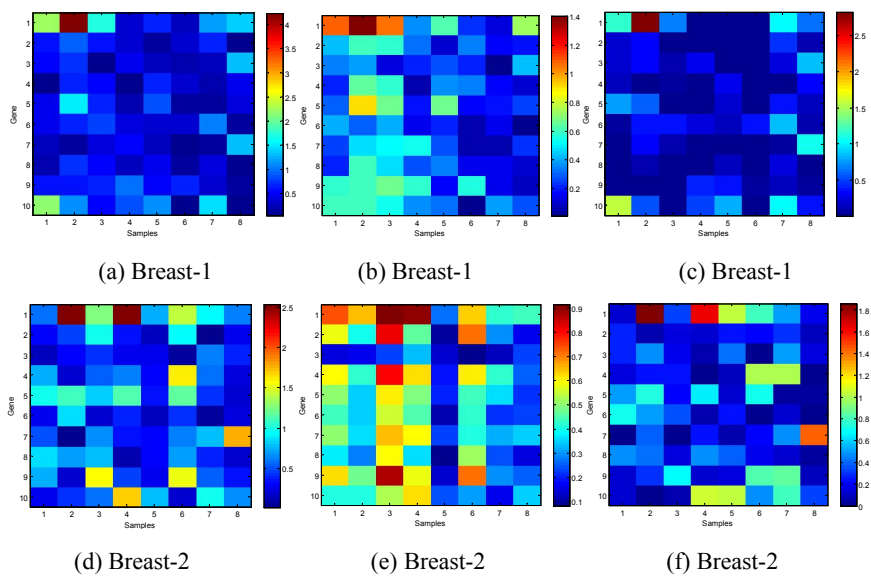

(a) Breast-1    (b) Breast-1    (c) Breast-1

(d) Breast-2    (e) Breast-2    (f) Breast-2

Fig.3. Pseudo color image of the samples for the Breast-2 database, (a), (d) observation matrix heat map, (b), (e) low-rank matrix heat map. (c), (f) variant dictionary matrix heat map.

In this experiment, for the breast tumor gene expression datasets, randomly select some of the two categories of samples to put together for low rank decomposition, we use the decomposition of the sparse part vector to construct a new variation dictionary. Fig.3 mainly describes the sparsity of variation dictionary from the heat map. (a), (d) observation matrix heat map； (b), (e) low-rank matrix heat map; (c), (f) variant dictionary matrix heat map; where transition from blue to red colors corresponds to a shift from low to high expression values of the samples. The vertical axis represents samples (4 tumor samples, 4 normal samples) and the horizontal the genes arranged. We use a sparsely constructed variation dictionary instead of the original gene expression profile data. Variation dictionary can be considered as ones of differential expression.

Matrix $D = A_0 + E_0$, analytic solution $(A, E)$, real solution $(A_0, E_0)$, $\lambda = 0.0707$.

Table.2 The error, rank and sparseness of the output solution and the real solution in Breast-1 data.

| iter | $\|\| A - A_0 \|\|_F$ | Rank($L$) | $\|\| E \|\|_0$ |
| --- | --- | --- | --- |
| 1 | 0.069465 | 69 | 3903 |
| 10 | 0.004144 | 66 | 7305 |
| 20 | 0.002572 | 54 | 8940 |
| 30 | 0.001442 | 48 | 9571 |
| 40 | 0.000851 | 44 | 9832 |
| 50 | 0.000484 | 43 | 9959 |
| 60 | 0.000298 | 41 | 10043 |
| 70 | 0.000210 | 41 | 10085 |
| 80 | 0.000165 | 40 | 10107 |
| 90 | 0.000132 | 40 | 10124 |
| 100 | 0.000107 | 40 | 10123 |

As can be seen from the experimental results, as the number of iterations increases, the error and rank become smaller and smaller, and the output solution becomes more and more sparse.

**3.2.2 Comparison of variation dictionary and other dictionary**

This section mainly introduces the contrast between our variation dictionary and ESRC[9] average sample variation dictionary, IPGSRC comparison of three classification methods. The Breast-2 datasets, our method first screened for 200 genes with SNR[22]. The data is then equally divided into ten categories based on the class labels, of which nine are put together to construct a variation dictionary using low-rank decomposition and the other one is put together for low-rank decomposition and the sparse fraction is used as a test.

Table.3 Comparison of different dictionaries of breast tumor.

| Method | SRC | GSRC | IPRC | IPGSRC |
|---|---|---|---|---|
| Our method | 66.67% | 77.78% | 77.78% | 88.89% |
| Average dictionary | 55.56% | 55.56% | 55.56% | 66.67% |

Table.4 Sensitivity and specificity of different classifiers.

| Method | SRC | GSRC | IPRC | IPGSRC |
|---|---|---|---|---|
| Sensitivity | 50% | 50% | 75% | 75% |
| Specificity | 80% | 100% | 80% | 100% |
| AUC | 0.90 | 1 | 0.65 | 1 |

From Table.3 and Table.4, we can see from the recognition rate, the sensitivity and specificity that the our method is better.

The current clinical manifestations of breast tumor, the specificity is relatively low, that is, the misdiagnosis rate is relatively high, the breast tumor misdiagnosed as breast hyperplasia, breast fibroids are still recurring, delayed treatment, seriously affecting the patient's survival time. Breast tumor misdiagnosed line local excision may have adverse consequences and all malignant tumors, the first time the correct treatment of local breast tumor and reasonable treatment is an important part of good effect, inappropriate local treatment may bring the some patients Dangerous. Visible high misdiagnosis, then the cost is relatively large, and our specificity of this method to 100%, that is misdiagnosis rate of 0.

## 3.3 Results of breast tumor classification based on variation dictionary and IPGSR

This part focuses on description IPGSRC convergence analysis, the performance of IPGSRC is also compared with the latest method based sparse representation for breast tumor classification. The compared methods are IPSRC, GSRC, SRC.

### 3.3.1 Convergence analysis

In this section, the convergence of IPGSRC model is analyzed. Fig.4 represent algorithms ADMM iterative error chart for solving IPGSRC model, where Fig.4 (a) the iteration error chart between exact solution and iterative solution, Fig.4 (b) the iteration error chart between adjacent iterations. We can be seen from the Fig.4 that the overall trend of the error of the algorithm is decreasing and tends to zero, indicating that the algorithm of ADMM is convergence.

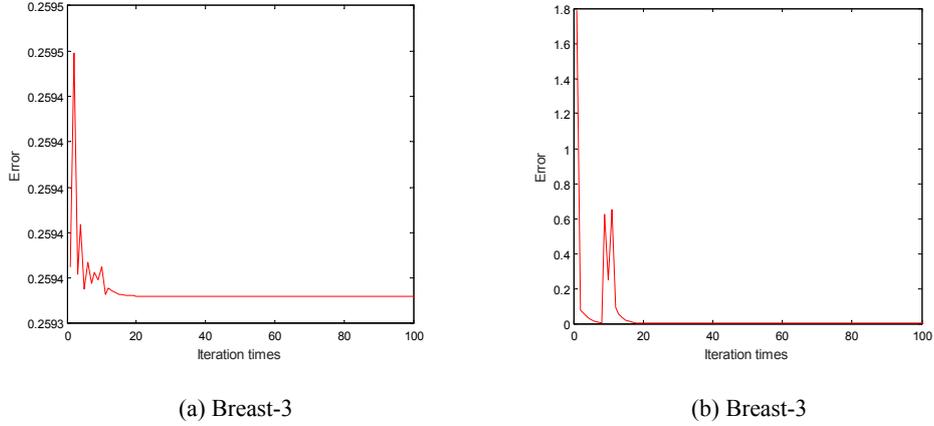

(a) Breast-3　　　　　　　　　　　　(b) Breast-3

Fig.4. Convergence analysis IPGSRC model. (a) IPGSRC the iteration error chart. (b) The error between two adjacent iterative solutions.

### 3.3.2 Comparison of IPGSRC and other classification

The performance of IPGSRC for robust breast tumor classification is demonstrated in this subsection. For comparison, the results of IPRC 、GSRC and SRC are listed under the same experimental environment.

The classification results, accuracy, sensitivity, specificity and AUC are listed in Table 5. For the two binary datasets, ROCs are drawn to further study the trade off between sensitivity and specificity of the approaches (Fig.5). For each experiment, we run the ten-fold cross validation ten times and take the means as the final results. Table.5 and Fig.5 show that IPGSRC achieves competitive results with highest AUC, which shows IPGSRC has the best prediction ability among the three classifiers. ROC plot analysis in Fig.5 has shown that IPGSRC has the better discrimination ability than IPRC、GSRC and SRC. The accuracy of IPGSRC are higher than IPRC、GSRC and SRC on Breast-2 dataset.

More intuitively, ERR is introduced to demonstrate that our method is superior to latest articles.

$$ERR = \frac{ER_1 - ER_2}{ER_1} \times 100\%,$$

where $ER_1$ is the error rate of the latest article classification result, $ER_2$ is the error rate of our method classification result, ERR is denoted by a notion ↓. Table.5 lists the error reduction results by switching SRC to IPRC on Breast-2 datasets. Table.5 show that since the IPGRC reduces the error rate from 29.56% to 11.37%, the ERR is 61.54% [(29.56-11.37)/29.56], suggesting that 61.54% recognition errors can be avoided by using IPSRC instead of SRC.

Table.5 Breast-2 datasets.

| Method | Accuracy | Error rate | ERR |
| --- | --- | --- | --- |
| IPGSRC+300 | 88.63% | 11.37 | - |
| IPRC+300 | 73.21% | 26.79 | ↓57.55% |
| GSRC+300 | 71.90% | 28.10 | ↓59.54% |
| SRC+300 | 70.44% | 29.56 | ↓61.54% |
| IPGSRC+1000 | 91.49% | 8.51 | - |
| IPRC+1000 | 88.02% | 11.98 | ↓28.96% |
| GSRC+1000 | 85.24% | 14.76 | ↓42.34% |
| SRC+1000 | 84.13% | 15.87 | ↓46.38% |

Table.6 Breast-3 data identification rate 100 average.

| methods | IPRC+200 | IPRC+200 | IPGSRC+200 | GSRC+200 | SRC+200 |
| --- | --- | --- | --- | --- | --- |
| accuracy | 99.75% | 99.67% | 100% | 99.67% | 99.33% |

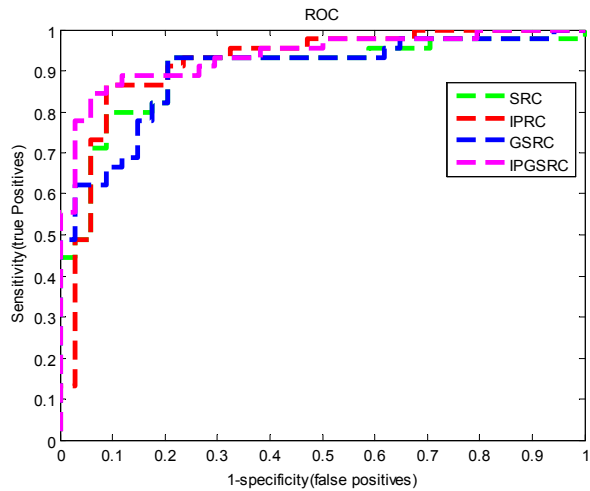

Fig.5. ROC analysis of the ability of SRC, GSRC, IPRC and IPGSRC on Breast-2 datasets. Note that on the vertical axis, the scale is from no (0) to complete (1 or 100%) sensitivity. The horizontal axis is a reciprocal scale (1-specificity). The optimum performance of a test is determined either as the highest sum of the specificity and sensitivity or at an acceptable level of sensitivity for the given disease.

Table.7 The AUC area corresponding to the ROC curve of different classifiers.

| Method | IPRC | SRC | GSRC | IPGSRC |
|--------|------|------|------|--------|
| AUC | 91.63 | 90.00 | 89.67 | 0.9353 |

Box plots of error rates are shown in Fig.6 after performing ten-fold cross validation. Fig.6 illustrates that SRC, GSRC, IPRC and IPGSRC achieve average error rates (red line) of 27%, 25.5%, 25.9%, 22.3% on Breast-2 datasets. Overall, a robust result with relatively low error rate can be offered by representation based methods.

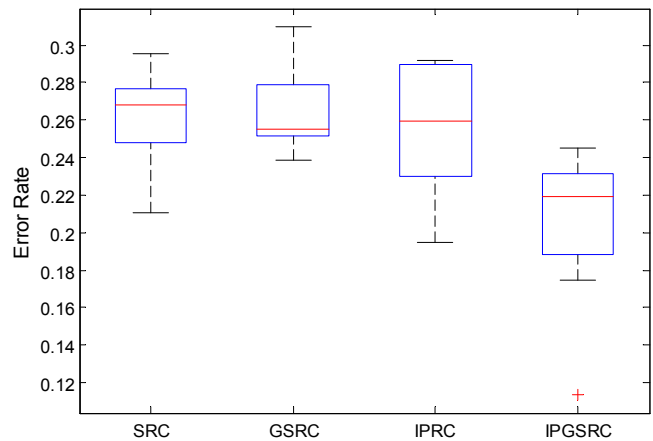

Fig.6. Box plots for error rates of four classifiers, The box characterizes the distribution of error rate. Namely, box's location and scale indicate the median (red lines) and interquartile range (blue box), respectively.

### 3.3.3 Comparison of Category Contribution Rate and Reconstruction Error

Next, we demonstrate CCR for IPRC is superior to reconstruction error for SRC. First, we should notice the fact that the more obvious the difference between categories is, the stronger the discrimination ability is, and the better the classification criterion is. Figs. 7 and 8 give the results of the two criterions about some randomly selected test samples. The same color expresses the values of a test sample across all categories. According to the overall trend, one can see that, to the same test sample, difference between categories of CCR is much bigger than that of reconstruction error. This shows that the CCR has better discrimination power than construction error.

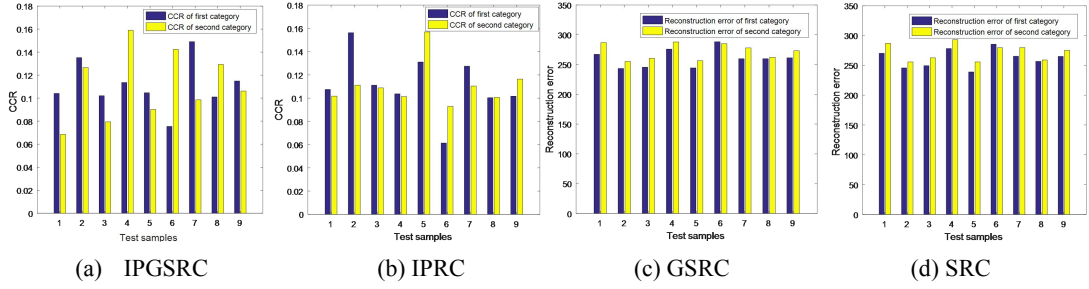

(a) IPGSRC  (b) IPRC  (c) GSRC  (d) SRC

Fig.7. The comparison of nine samples random selected on Breast-2 datasets. (a),(b) CCR, (c),(d) Reconstruction error. The same color histogram expresses the same category.

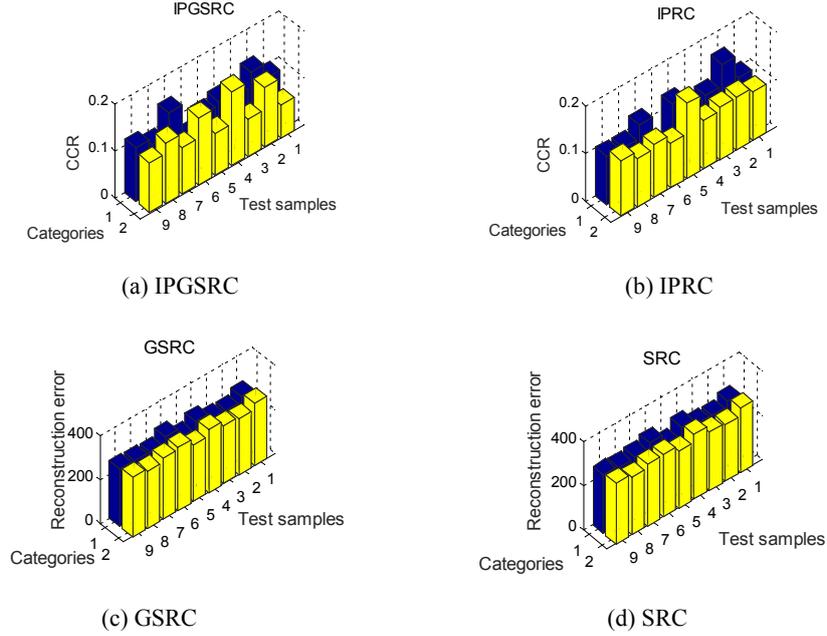

(a) IPGSRC  (b) IPRC

(c) GSRC  (d) SRC

Fig.8. The comparison on Breast_2 dataset. (a),(b) CCR. (c),(d) Reconstruction error. This 3-dimensional histogram shows the values of two classification criterions of test samples across all classes. The same color expresses the values of the same test sample across all categories.

### 3.3.4 Results of classification stability

Next the classification stability is further verified by comparing the quantitative indicator of stability, CSI. Fig.9 shows the CSI of all samples on the Breast-2 datasets. The smaller the CSI is, the better the stability is. One can see that $CSI_{RE}$ is almost close to 1 in all subfigures, while the $CSI_{CCR}$ is much smaller. That is to say, the difference between $R_{best}^1$ and $R_{best}^2$ in CCR for IPGSRC is much bigger than those in reconstruction error for SRC. This further verifies CCR for IPGSRC has better.

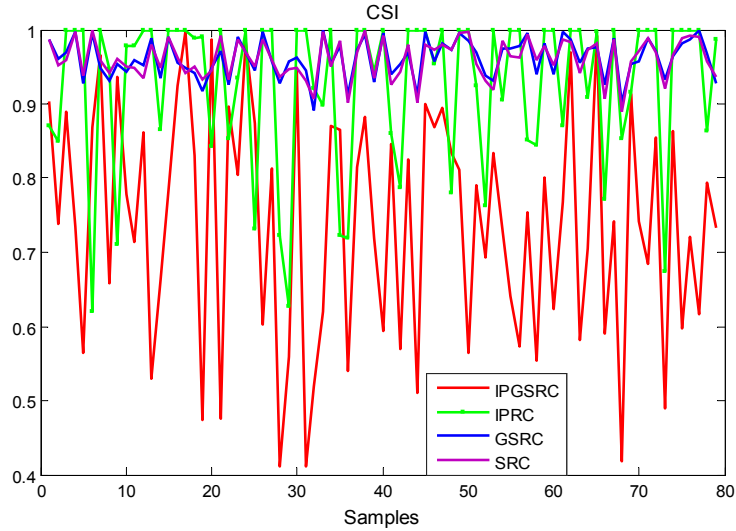

Fig.9. The curve of CSI versus all samples in ten-fold cross validation. The red line expresses CSIs of CCR in IPGSRC, green line expresses CSIs of CCR in IPRC, blue line expresses CSIs of reconstruction error in SRC+G, and purple line expresses CSIs reconstruction error in SRC. The smaller the CSI is, the more stable the representation is.

### 3.3.5 Comparing with state-of-the-art methods

This part explores the classification results of breast tumors based on the Breast-2(97), Breast-1database in recent years compared with the classification results of this article. Table 8 shows that on the Breast-2 (97) dataset, the latest articles in recent years use 10-fold cross validation, the highest accuracy was 87.4%. Our method in the same dataset and same environment, the accuracy was 87.1%. Table 9shows that on the Breast-1 dataset, the latest articles in recent years the highest accuracy was 80.03%. Our method in the same dataset and same environment, the accuracy was 80.81%, 0.78% above the highest accuracy.

Table 8 and 9 show that our method classification performance is better than latest classification results. For Breast-2(97) dataset, although the results of the method are higher than ours, Su et al. still start from the perspective of information genes, we use new thinking, from the viewpoint of detecting and using variations in gene expression profiles of normal and patients, rather than directly using these samples.

Table.8 Classification accuracies based on different methods on Breast-2(97).

| Methods | Accuracy(%) |
|---|---|
| Jiang et al.(2017) | 80 |
| Su et al.(2017) | 87.4 |
| Our method | 87.1 |

Table.9 Classification accuracies based on different methods on Breast-1.

| Methods | Error Rate（%） | ERR(%) |
|---|---|---|
| Our method | 19.19 | - |
| Zhang et al.(2017) | 20 | ↓4.05 |
| Fan et al.(2015) | 19.97 | ↓3.91 |

## 4 Analysis of candidate's pathogenic genes

Apart from obtaining high classification accuracy results, it is also important to identify pathogenicity-related genes, which can be a biomarker of early diagnosis and be helpful to auxiliary diagnosis. Pathogenicity-related genes selected in this article are decision information genes. Firstly, the basic biological attribute description of the decision information genes for classification is given, and then the specific biological description and rationality verification of some selected decision information genes are given.

### 4.1 Gene biology analysis

### 4.1.1 Gene enrichment analysis

To further study the biological function of the candidate pathogenicity-related genes. We first selected 300 genes from the SNR gene. We also perform the functional enrichment analysis of the top 300 genes identified by our method on the website https://david.ncifcrf.gov/. The results of KEEG_PATHWAY are listed in Table 10. It can be seen from Table.10 that the item of metabolic pathways has the lowest p-value, so it is considered as the most probable enrichment item. Only this pathway has statistical meaning( $p<0.05$ ). For genes enrich by our method from Breast-2 data, we further do Kaplan-Meier curve by anglicizing survival curves and corresponding Log-Rank p values on the website http://www.oncolnc.org and http://ualcan.path.uab.edu/index.html. It can be seen from this Table.11 that there are six genes with statistical meaning (p<0.05), five are proto-oncogenes and one anti-oncogene. On Fig.10 we show the survival curve of 2 genes. From Fig.13, we can know that PGK1 (p=2.58e-06) is anti-oncogene, IP6K2 (p=0.0244) is proto-oncogenes. For the two genes, high expression and low expression have significant difference in survival rate.

Similar to Breast-2 data, on Breast-3 data used SNR for preliminary gene screening of 200 genes. We also perform the functional enrichment analysis of the top 200 genes identified by our method. The results of KEEG_PATHWAY are 55 genes (P <0.05 in the gene pathway). For genes enrich by our method from Breast-3 data, we further do Kaplan-Meier curve by anglicizing survival curves and corresponding Log-Rank p values. It can be seen from this Table.12 that there are 2 are proto-oncogenes. On Fig.11 we show the survival curve of two genes.

Table.10 KEEG-PATHWAY terms enrichment analysis of the top 300 genes in the Breast-2 data set by DAVID

| Rank | KEGG_PATHWAY | P-Value |
|------|--------------|---------|
| 1 | Pyrimidine metabolism | 6.9e-3 |
| 2 | Oocyte meiosis | 8.4e-3 |
| 3 | Metabolic pathways | 1.0e-2 |
| 4 | Cell cycle | 1.4e-2 |
| 5 | Glutathione metabolism | 2.1e-2 |
| 6 | Phosphatidylinositol signaling system | 2.6e-2 |
| 7 | p53 signaling pathway | 4.2e-2 |
| 8 | Purine metabolism | 5.3e-2 |
| 9 | Progesterone-mediated oocyte maturation | 7.9e-2 |
| 10 | Fatty acid degradation | 9.2e-2 |

Table.11 Proto-oncogenes and anti-oncogene searched by TCGA

| Gene name | P-Value | Gene | Genebank_Accession |
|---|---|---|---|
| CTPS1 | 0.046 | Proto-oncogene | 1503 |
| GMPS | 0.0215 | Proto-oncogene | 8833 |
| PGK1 | 2.58E-06 | Proto-oncogene | 5230 |
| PGAM1 | 0.0168 | Proto-oncogene | 5223 |
| CCNB2 | 0.0376 | Proto-oncogene | 9133 |
| IP6K2 | 0.0244 | anti-oncogene | 51447 |

Table.12 Proto-oncogenes and anti-oncogene searched the 26 genes in the Breast-3 data set by TCGA

| Gene name | Genebank_Accession | P-Value | Gene |
|---|---|---|---|
| SFRP1 | 6422 | 0.00486 | Proto-oncogene |
| NTRK2 | 4915 | 0.00796 | Proto-oncogene |

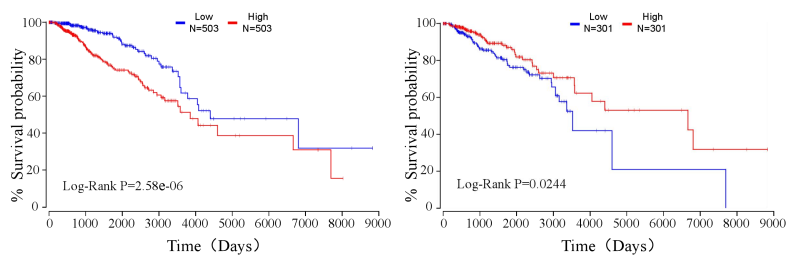

Fig.10. Kaplan-Meier survival curves of genes which enrich in pathways with statistical mearning ($p < 0.05$). Subimages from left to right: PGK1 and IP6K2 respectively.

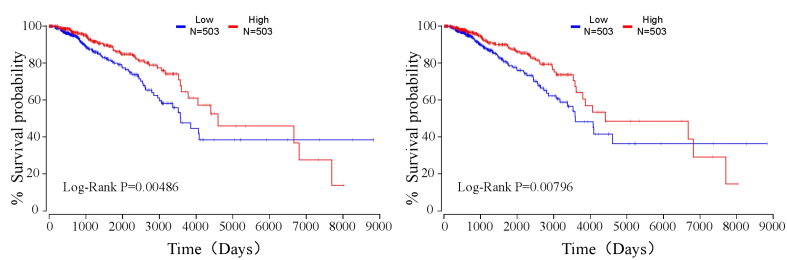

Fig.11. Kaplan-Meier survival curves of the genes in the Breast-3 data set by TCGA which enrich in pathways with statistical meaning ($p < 0.05$). Subimages from left to right: SFRP1 and NTRK2 respectively.  Red lines denote upper 50% percentile and blue lines denote lower 50% percentile.

### 4.1.2 Analysis of candidate's pathogenic genes

In order to check the quality of the selection processes, the expression profiles of the final identified genes for the opposite category are analyzed. For comparison, two irrelevant genes chosen randomly are presented. Fig.12 illustrates the two exemplary expression levels of the patients for the pathogenic gene (CTPS1、GMPS、PGK1、PGAM1、CCNB2 and IP6K2) listed in Table 11 and four irrelevant genes (NM_014670andContig25583_RC). In Fig.12, the red line denotes gene expression levels of 45 normal samples and the blue line expresses gene expression levels of 34tumor samples. The line indicates the mean values of gene expression levels in corresponding class. One can see in both cases the mean value of the samples belonging to breast tumor category differs significantly from the referenced (normal) category.

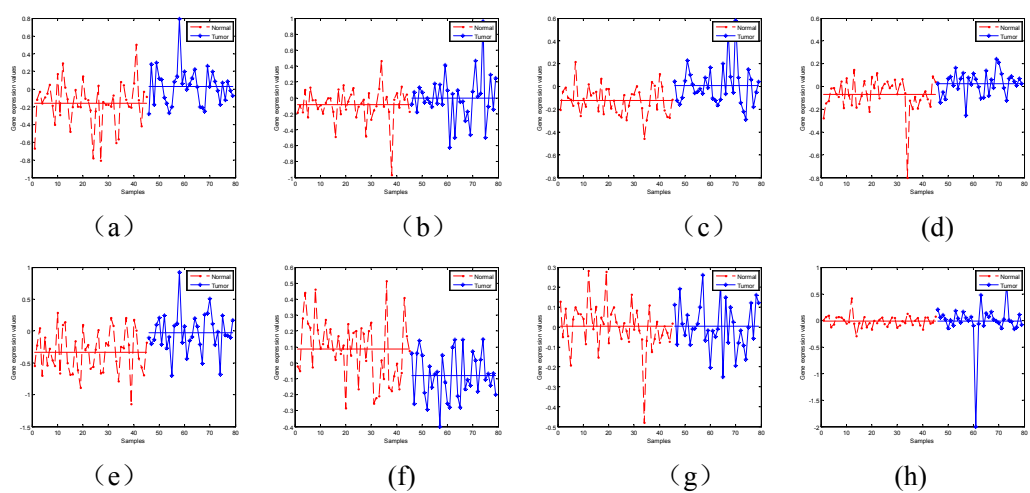

（a） （b） （c） （d）

（e） （f） （g） （h）

Fig.12.Comparison of expression levels for the pathogenic genes (a, b, c, d, e, f, g, h) and irrelevant genes (g, h). For pathogenic genesCTPS1, GMPS, PGK1, PGAM1, CCNB2 and IP6K2, the mean and standard deviation of expression levels about tumor category samples are higher than that of normal category samples. But for irrelevant genes NM_014670 and Contig25583_RC, they have similar mean and std in both categories.

Further more vividly illustrate the results of this article, Fig.14 shows the image of the expression profiles for the two pathogenic genes (CCNB2 and IP6K2) and two irrelative genes in the form of the colormap of jet, where transition from blue to red colors corresponds to a shift from low to high expression values of the samples. The vertical axis represents samples. Fig.12 and Fig.13 demonstrate that moderate to high up regulation of TSPYL5 and ATP5J and down regulation for other two genes can indicate the presence of Breast tumor distant metastases. There is a visible difference between samples of the Breast tumor group and the reference one in TSPYL5 and ATP5J but similar expression levels in PTPN1 and ATP2C2-AS1, which confirms good performance of the proposed gene selection procedure.

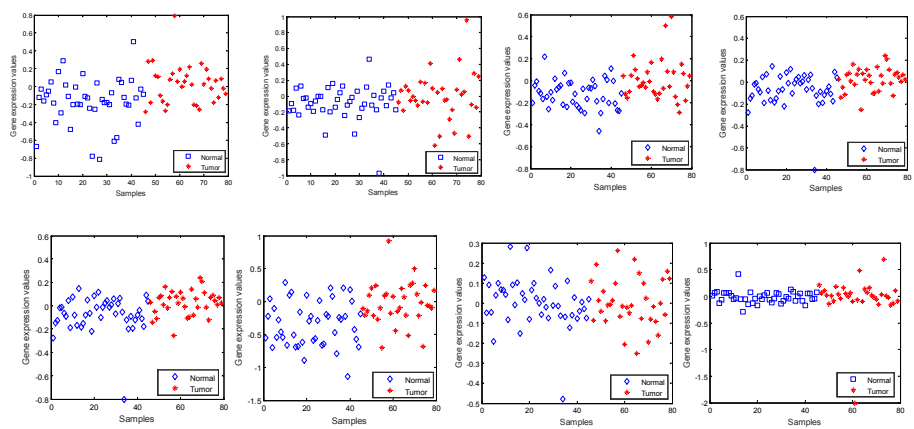

Fig.13 Scatter plots of two identified genes and two irrelative genes on Breast-2 datasets. Each panel corresponds to one gene. The red and blue colors correspond to normal and tumor types respectively. (a, b, c, d, e, f) pathogenic genes (CTPS1, GMPS, PGK1, PGAM1, CCNB2 and IP6K2), and (g, h) NM_014670 and Contig25583_RC are equally expressed for tumor and normal.

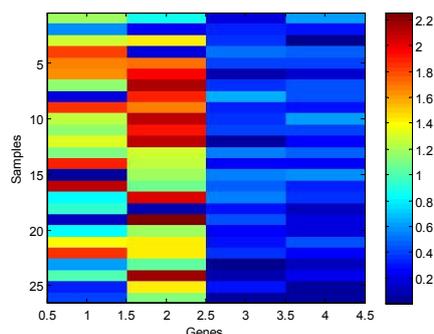

Fig.14.Pseudo color image of the samples for the Breast-2 datasets. Each panel corresponds to one gene. Blue to red colors represents low to high expression levels of samples. The image reveals that moderate to high up regulation of CCNB2 and IP6K2 and down regulation for other two genes can signal Breast-2.

**4.2 Biological Description**

PGK1 biological description: one, Data suggest that, in breast tumor cells, MYC acts as an upstream regulator leading to PGK1 activation. (MYC, proto-oncogene c-myc; PGK1, phosphoglycerate kinase 1).Two, results show that PGK1 mRNA and protein expression were significantly increased in breast tumor tissues and can be considered as a prognostic biomarker of chemoresistance to paclitaxel treatment in breast tumor.

IP6K2 biological description: In yeast, IP6K plays a role in modulating telomere length, soHSP90:IP6K2 interaction may also be anew target for controlling telomerelength in tumor cells. The ability ofHSP90 to inhibit the normal growth regulatory activity of IP6K2 may be a segueinto the discovery of several similar interactions.

PGAM1 is an enzyme involved in glycolysis and may play a role in the metabolism and proliferation of tumor cells. PGAM1 is highly expressed in tumor cells. PGAM1 mRNA is expressed in both astrocytes and c6 glioma cells. Expression was significantly higher in C6 glioma cells than in astrocytes. It was hypothesized that PGAM1 mRNA plays a role in the malignant growth of gliomas. The expression of PGAM1 protein in gliomas is significantly higher. Peri-tumoral tissue further demonstrated that PGAM1 is associated with the malignant growth of gliomas. It fully suggests that PGAM1 is expected to become a new tumor marker and therapeutic target for gliomas.

CTPS1 biological description: Deficiency in cytidine 5-prime triphosphate synthase 1 (CTPS1) causes a combined immunodeficiency affecting both cellular and humoral immunity. CTPS1, one of two mammalian CTP synthases (the other being CTPS2), catalyzes the formation of CTP, an essential nucleotide for DNA and RNA synthesis . Lymphocytes from patients with CTPS1 deficiency have reduced intracellular levels of CTP, which suggests a defect in protein stability. The diagnosis of CTPS1 deficiency is difficult. Patients can have a relatively normal immunologic workup, especially in the absence of chronic EBV infection .

**5 Conclusions and future work**

In this paper, an inverse-projection group sparse representation model is presented for breast tumor classification, which is based on constructing low-rank variation dictionary. The proposed low-rank variation dictionary tackles tumor recognition problem from the viewpoint of detecting and using variations in gene expression profiles of normal and patients, rather than directly using these samples. The inverse projection group sparsity representation model is constructed based on taking full using of exist samples and group effect of microarray gene data.

Furthermore, some valuable analysis of candidate pathogenicity-related genes is given. There remain some interesting questions. One is how to enforce some prior constraints into the IPGRC model and construct gene network. Another is gene selection and its biological meanings.

**Acknowledgments**

The authors would like to thank https://tumorgenome.nih.gov/ for their breast datasets. We also thank Weifeng Yue for bioinformatics' suggestion, Chenxi Tian and Lei Sun for discussion. This work was supported in part by National Natural Science Foundation of China (11701144), Key Project of the Education Department Henan Province (14A120009), Natural Science Foundation of Henan Province (162300410061) and Project of Emerging Interdisciplinary (xxjc20170003). X.H. Yang is the corresponding author.